\documentclass[10pt,conference]{IEEEtran}
\IEEEoverridecommandlockouts

\usepackage{cite}
\usepackage{amsmath,amssymb,amsfonts}
\usepackage{algorithmic}
\usepackage{graphicx}
\usepackage{textcomp}
\usepackage{xcolor}
\usepackage[utf8]{inputenc}
\usepackage{textgreek}
\usepackage{multirow}
\usepackage{booktabs}
\def\BibTeX{{\rm B\kern-.05em{\sc i\kern-.025em b}\kern-.08em
    T\kern-.1667em\lower.7ex\hbox{E}\kern-.125emX}}
\begin{document}

\title{Multi-Scale Convolution with Optimal Transport Attention Effect on Multivariate Time Series\\
}

\author{\IEEEauthorblockN{1\textsuperscript{st} HaoChong Fu}
\IEEEauthorblockA{\textit{Institute of Collaborative Innovation} \\
\textit{University of Macau}\\
Macau, China \\
fantaisiedemickey@gmail.com}
\and
\IEEEauthorblockN{2\textsuperscript{nd} Jian Xu}\IEEEauthorblockA{\textit{}
\textit{RIKEN AIP}\\
Tokyo, Japan \\
jian.xu@riken.jp}
}

\maketitle

\begin{abstract}
The analysis of Multivariate Time Series (MTS) plays an important role in a lot of real-world practical applications, but it still remains some challenging problem about capturing multi-granularity structural patterns and suppressing noise appropriately. 
Multi-Scale Convolution with Optimal Transport Attention (MSC-OT) is proposed in this paper. MSC-OT is a useful architecture to optimize the attention mechanism. It combines multi-scale convolution with Sinkhorn optimal transport method based on inverted embedding. The inverted embedding approach embeds each variable as a token and allows the model to capture cross-variate relationships better. MSC-OT consists of two part: (1) Multi-Scale Convolution Enhancement, that applies multi-scale convolutions to attention score matrices based on inverted embedding, capturing local structural patterns in the variate-interaction space induced by compressed temporal representations;  (2) Sinkhorn Optimal Transport Regularization, that formulates attention computation as an optimal transport problem and employs iterative matrix scaling to ensure balanced information flow across variates.
Adaptive Fusion Strategy utilizes softmax-normalized learnable weights to dynamically combine base attention, convolution-enhanced, and OT-regularized scores. Experiments on widely-used datasets, including ETT, Electricity, Traffic, Solar-Energy, and Exchange-Rate, show that MSC-OT achieves well performance in both short-term and long-term forecasting tasks. Ablation experiments further validate the effectiveness of each proposed component and their synergistic contributions to improving prediction accuracy for multivariate time series forecasting.

\end{abstract}


\section{Introduction}

Transformer \cite{vaswani2017attention} has achieved a success in natural language processing \cite{brown2020language} and computer vision \cite{dosovitskiy2020image}, emerging as the dominant architecture for sequence modeling tasks. Inspired by this success, Transformer-based models have been extensively adapted for time series forecasting, with different type of innovations including sparse attention mechanisms \cite{zhou2021informer}, auto-correlation decomposition \cite{wu2021autoformer}, and frequency-enhanced representations \cite{zhou2022fedformer}.

But researchers have recently questioned the validity of conventional Transformer-based forecasters, which typically embed multiple variates of the same time into channels and apply attention on temporal tokens. Considering the numerical but less semantic relationship among time points, simple linear layers have been shown to exceed complicated Transformers on both performance and efficiency \cite{zeng2023transformers}. ITransformer \cite{liu2023itransformer} addressed it by introducing an inverted embedding paradigm that embeds entire time series of each variate as a single token. This type of embedding enable the attention mechanism to explicitly capture cross-variate dependencies rather than temporal dependencies. However, we observe that the inverted embedding paradigm raises a critical limitation: by embedding each variate's temporal information into a fixed-dimensional token, fine-grained temporal patterns important to prediction may be lost. 

\begin{figure}[htbp]
\centerline{\includegraphics[width=0.5\textwidth]{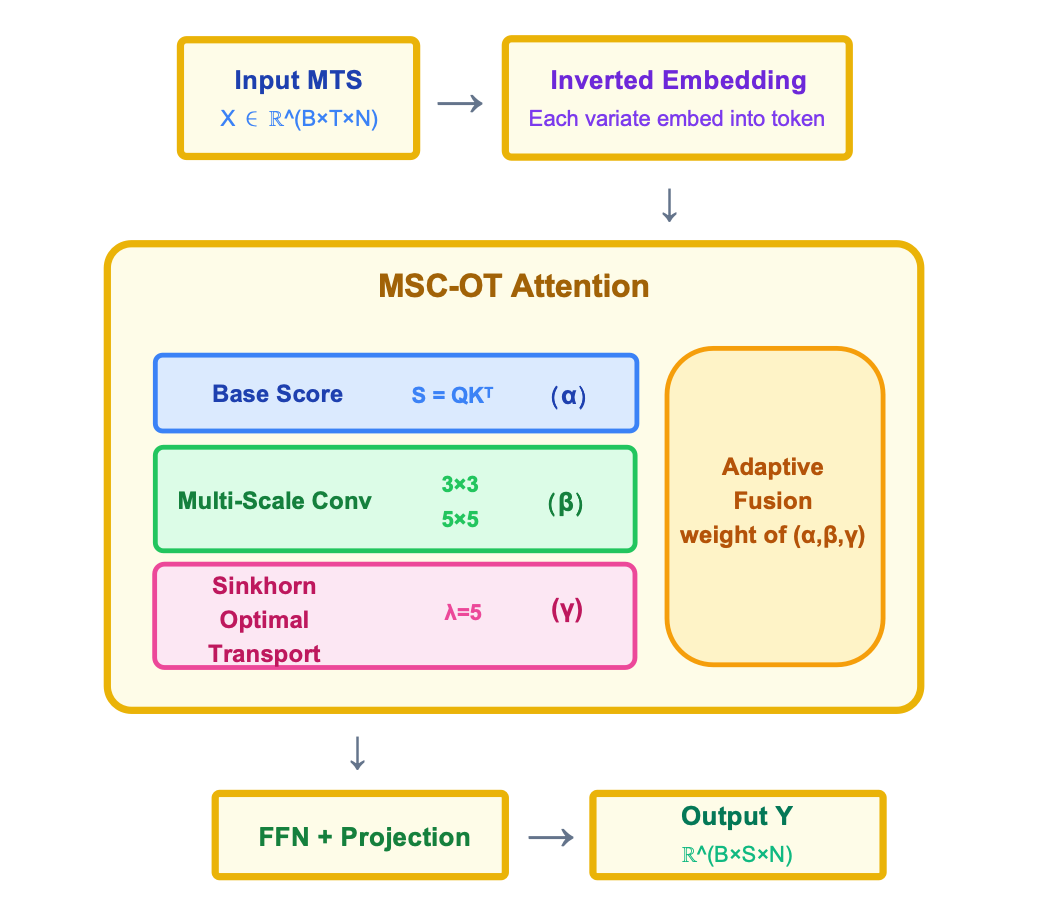}}
\caption{In MSC-OT, we use $\alpha$ to define the base score weighting, $\beta$ to define the Multi-Scale Convolution weighting, $\gamma$ to define the Sinkhorn Optimal Transport weighting, and the final result is calculated using an adaptive fusion algorithm.}
\label{fig}
\end{figure}

Considering these limitations, we focus on how to enhance the inverted embedding framework while preserving its advantages for cross-variate modeling. MSC-OT (Multi-Scale Convolution with Optimal Transport) Attention is proposed, an attention mechanism that addresses the inherent limitations. Technically, we apply parallel 2D convolutions with kernel sizes of $3\times3$ and $5\times5$ to the attention score matrix, capturing local structural patterns at different granularities that reflect compressed temporal information. We formulate attention computation as an entropy-regularized optimal transport problem and employ the Sinkhorn algorithm to achieve doubly-stochastic attention distributions that are inherently robust to anomalous noise. And we introduce an adaptive fusion strategy with learnable weights to dynamically combine base attention, convolution-enhanced scores, and OT-regularized scores based on five datasets. Experimental results show that MSC-OT performs well on five datasets, achieving an improvement compared to several time series models that have been widely cited in research over the past few years, especially compared to time series methods based on other embedding methods, showing a more significant improvement (improve $27\%$ compared to Crossformer that based on DSW and TSA; improve $35\%$ compared to TimesNet that transforming 1D time series into 2D tensors; improve $19\%$ compared to Fedformer that based on frequency-domain sparse representations). The overall structure of MSC-OT is shown in Fig. 1.

Note: The code available has been uploaded in GitHub FantaisieDeMickey/msc-ot/.

\section{Related Works}
\subsection{Deep Learning for Multivariate Time Series Forecasting}
Multivariate time series (MTS) forecasting has evolved significantly with the development of deep learning. Early neural approaches employed TCNs \cite{bai2018empirical} and RNN-based models \cite{salinas2020deepar}\cite{rangapuram2018deep} to capture temporal dependencies, treating MTS data as sequences of vectors. LSTnet \cite{lai2018modeling} combined CNN for cross-dimension dependency and RNN for cross-time dependency. Graph neural network-based methods \cite{li2017diffusion}\cite{wu2020connecting} explicitly modeled cross-variate relationships through graph convolution. However, these models face difficulty in capturing long-term dependencies due to the inherent limitations of CNN and RNN architectures.

Transformer-based models have emerged as a dominant paradigm for MTS forecasting. Informer \cite{zhou2021informer} introduced ProbSparse attention to reduce complexity from $O(L^2)$ to $O(L log L)$. Autoformer \cite{wu2021autoformer} proposed series decomposition with auto-correlation mechanisms. FEDformer \cite{zhou2022fedformer} leveraged frequency-domain sparse representations. Pyraformer \cite{liu2022pyraformer} developed pyramidal attention for multi-resolution temporal modeling. However, these methods embed multiple variates at the same timestamp into a single token, focusing primarily on temporal dependency while inadequately capturing cross-variate correlations.

Recent works have shifted focus toward explicit multivariate modeling. Crossformer \cite{zhang2023crossformer} proposed Dimension-Segment-Wise embedding and Two-Stage Attention to capture both cross-time and cross-dimension dependencies. PatchTST \cite{nie2022time} adopted channel-independent patching but overlooked inter-variable dependencies. ITransformer \cite{liu2023itransformer} introduced the inverted embedding paradigm that treats each variate's entire time series as a token, enabling attention to model cross-variate correlations directly. Some models also use novel methods for time series prediction, such as D2Vformer \cite{song2025d2vformer} which devises to leverage timestamp information and feature sequences to generate time-position embeddings, and TimeCMA \cite{liu2025timecma} which integrates LLM for time series analysis. Our MSC-OT builds upon iTransformer's inverted framework while addressing its limitations in temporal dependency capturing and noise robustness through multi-scale convolution enhancement and optimal transport regularization.

\subsection{Multi-Scale Convolution for Time Series}
SCINet \cite{liu2022scinet} exploits the property that temporal relations are largely preserved after downsampling, proposing a recursive downsample-convolve-interact architecture. TimesNet \cite{wu2022timesnet} takes a different approach by transforming 1D time series into 2D tensors based on discovered periodicities. MICN \cite{wang2023micn} addresses both local feature extraction and global correlation modeling through a multi-scale branch structure. More recently, MSDCN \cite{li2024multi} proposes shallow multi-scale dilated convolution modules with exponentially growing dilation rates and varying kernel sizes to sample time series at different temporal scales. TimeMixer \cite{wang2024timemixer} proposes a decomposable multiscale mixing architecture that separately processes seasonal and trend components at different temporal scales through MLP-based mixing blocks. ModernTCN \cite{luo2024moderntcn} redesigns the time-convolutional-networks(TCN) and shows that large-kernel convolutions can effectively capture long-range temporal dependencies while maintaining computational efficiency. FITS \cite{xu2024step} shows that the method of simple linear interpolation on frequency can take a good forecasting performance.

\subsection{Sinkhorn Optimal Transport in Machine Learning}
Optimal transport (OT) provides a principled mathematical framework for comparing probability distributions. Cuturi introduced entropic regularization enabling efficient computation through the Sinkhorn-Knopp algorithm \cite{cuturi2013sinkhorn}, which alternates between row and column normalizations to converge orders of magnitude faster than linear programming solvers while maintaining full differentiability. This breakthrough has enabled OT integration into deep learning architectures.

Several works have applied OT principles to attention mechanisms. Sparse Sinkhorn Attention was proposed in employing meta sorting networks to learn latent permutations \cite{tay2020sparse}, reducing memory complexity through Causal Sinkhorn Balancing and SortCut for dynamic sequence truncation. Sinkformers was introduced in replacing softmax with Sinkhorn normalization to produce doubly-stochastic attention matrices \cite{sander2022sinkformers}. Gumbel-Sinkhorn Networks was developed for learning latent permutations through temperature-controlled Sinkhorn operators with Gumbel noise injection \cite{mena2018learning}.

Unlike these works that replace the entire attention computation with OT-based formulations, MSC-OT employs Sinkhorn OT as a complementary regularization component that is adaptively fused with base attention and convolution-enhanced scores. This design preserves the discriminative capacity of standard attention while leveraging OT's inherent denoising capability—the doubly-stochastic constraint prevents any single variate from dominating attention even with anomalously high scores, and the entropy regularization naturally suppresses concentrated distributions sensitive to outliers.

\section{MSC-OT: Multi-Scale Convolution with Optimal Transport Attention}
This section introduces our proposed MSC-OT (Multi-Scale Convolution-Optimal Transport) attention method, which aims to address the limitation of weak local temporal patterns capture capability in inverted embedding strategies in multivariate time series forecasting and achieve better results in noise suppression. We first present a problem description, then detail the various components of the architecture, and finally provide the adaptive score fusion.

\subsection{Problem Formulation}
Given a multivariate time series dataset $X \in \mathbb{R}^{B \times T \times N}$, the goal of long-term time series forecasting is to predict the next S time steps: $Y \in \mathbb{R}^{B \times S \times N}$ . Unlike the traditional Transformer architecture, which embeds time unit into all variables, the iTransformer introduces an inverted embedding paradigm that treats the entire time series of each variable as a single token, resulting in an embedding representation $H \in \mathbb{R}^{B \times N \times D}$, where D is the embedding dimension.The inverted embedding does not model temporal dependencies any more,  which may lead to the loss of fine-grained temporal local patterns for accurate prediction.

\subsection{Architecture Overview}
To address the limitations, we propose MSC-OT, a novel attention mechanism that integrates two complementary components: (1) Multi-Scale Convolution Enhancement for capturing multi-granularity structural patterns within the attention score matrix, and (2) Sinkhorn Optimal Transport for robust attention weight computation with noise suppression. 

\begin{figure}[htbp]
\centerline{\includegraphics[width=0.5\textwidth]{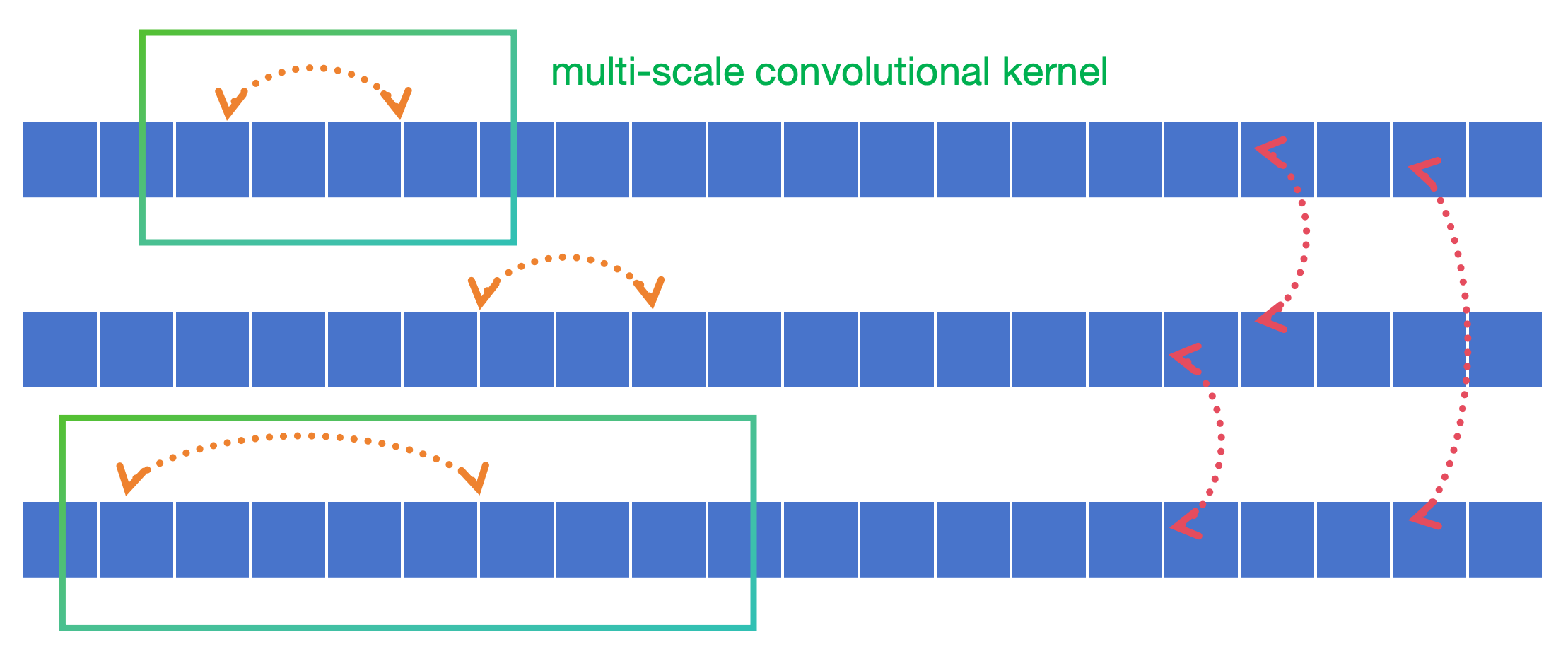}}
\caption{Inverted embedding is kept for capturing the variate dependency, and the multi-scale convolution is used to compensate for time correlations.}
\label{fig}
\end{figure}

\subsection{Multi-Scale Convolution Enhancement}

Although iTransformer takes a significant result by inverted embedding method, the limitation of capturing local structure dependency is idendified. In the iTransformer embedding structure, the time series of each variate $\mathbf{X}_{:,n} \in \mathbb{R}^{T}$ is projected into a D-dimensional variate token by a linear transformation:
\begin{equation}
\mathbf{h}_n = \mathbf{X}_{:,n} \mathbf{W}_\text{embed} + \mathbf{b}_\text{embed} , \quad \mathbf{W}_\text{embed}  \in \mathbb{R}^{T \times D}
\end{equation}

The linear projection operates as a global receptive field over the entire temporal dimension and good at capture holistic series characteristics (e.g., overall trend, mean amplitude). But it embeds all temporal positions of each variate in a token, and lacks explicit mechanisms to model local temporal structures at multiple scales. But the short-term fluctuations, periodic patterns, and hierarchical temporal dependencies are cannot be ignored.

Addtionally, although the feed-forward network (FFN) in iTransformer is applied to each variate token to learn nonlinear temporal representations, its fully-connected nature still operates on a globally-compressed representation, limiting its ability to preserve and exploit fine-grained temporal dynamics. This architectural choice implicitly assumes that all relevant temporal information can be adequately captured through linear compression followed by nonlinear transformation—an assumption that may not hold for time series exhibiting complex multi-scale temporal structures.

Our process is that temporal dynamics within each variate should be explicitly modeled. We argue that the quality of multivariate correlation learning in the attention mechanism is fundamentally dependent on the richness of the individual variate representations. If critical local temporal patterns are lost during embedding, the subsequent attention mechanism—regardless of its sophistication—cannot recover this information.
Integrating multi-scale convolutional operations within the attention mechanism is proposed to solve this kind of limitation.
Given the raw attention scores $S = \frac{QK^T}{\sqrt{d_k}} \in \mathbb{R}^{B \times H \times N \times N}$, where $B$ is the batch size, $H$ is the number of attention heads, and $N$ is the number of variates, we apply a set of $K$ convolutional filters with different kernel sizes {$k_1$, $k_2$, ..., $k_K$}:
\begin{equation}
S^{(i)}_{\text{conv}} = \text{Conv2d}(S; k_i), \quad i = 1, 2, \ldots, K
\end{equation}
where Conv2d(·; k) denotes a 2D convolution operation with $k \times k$ size kernel. In our experiments, we set $K = 2$ scales with kernel sizes 3 and 5, designed to capture  multi-granularity structural patterns interactions.
The multi-scale features are then aggregated through simple averaging:
\begin{equation}
S_{\text{conv}} = \frac{1}{K} \sum_{i=1}^{K} S^{(i)}_{\text{conv}}
\end{equation}
This averaging strategy ensures numerical stability while preserving the complementary information captured at different scales. The convolution weights are initialized with small constant values to prevent disrupting the original attention patterns during early training.

\subsection{Sinkhorn Optimal Transport on Attention}
 A single outlier can significantly influence the attention distribution, which means standard softmax attention are susceptible to outliers and noise in the input data. This phenomenon is particularly obvious in MTS. The data commonly used in MTS experiments usually contains measurement noise, missing values, and special points. It maybe cause the model to overemphasize special features and making the result offset to expected. 

\begin{figure}[htbp]
\centerline{\includegraphics[width=0.5\textwidth]{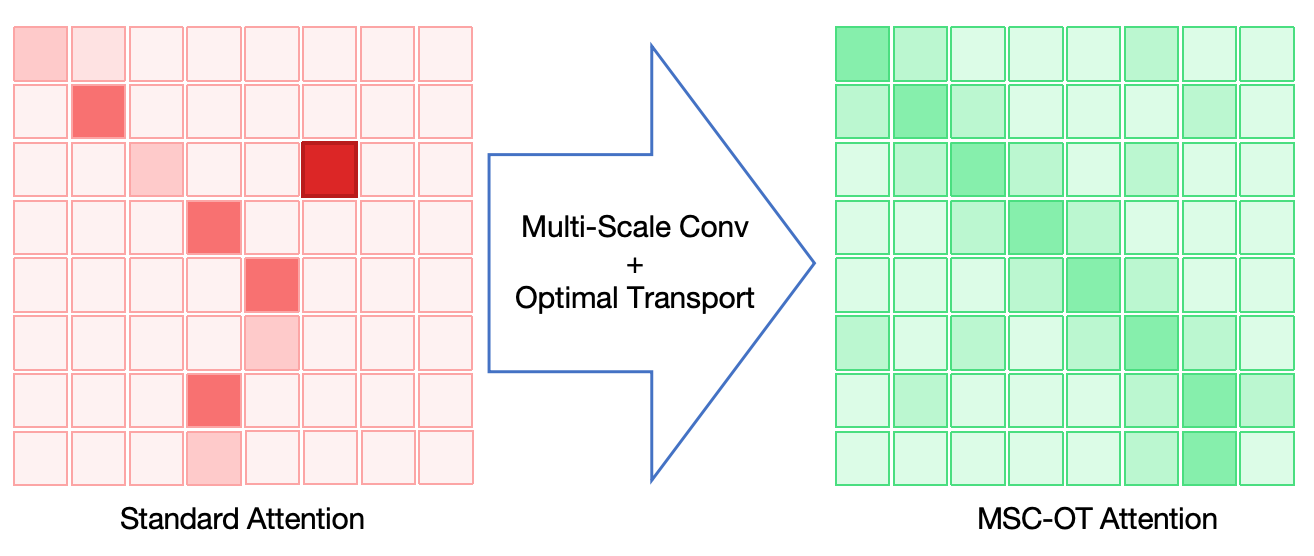}}
\caption{The optimal transport integrated into the attention mechanism (efficiently solved through the Sinkhorn algorithm) suppresses the influence of outliers, ensuring the rationality of the attention distribution.}
\label{fig}
\end{figure}

Optimal Transport (OT) theory offers a simple method to solve this problem. It makes the model more focused on the overall distribution and partially ignoring outliers by transforming the attention mechanism to a optimal transport problem. It is aimed to find an allocation matrix that minimizes the cost of "transferring" the query to the key while satisfying marginal constraints. The entropy regularization term in the Sinkhorn algorithm, an optimal transport algorithm, naturally smooths the final attention weights, effectively suppressing the impact of outliers.

We formulate the attention computation as an entropy-regularized optimal transport problem. Given the attention score matrix $S$, we compute the transport plan $P$ by solving:
\begin{equation}
P^* = \arg\min_{P \in \mathcal{U}(r,c)} \langle C, P \rangle - \lambda H(P)
\end{equation}
where $C = -S$ is the cost matrix (negative scores represent low transportation cost), $\mathcal{U}(r, c) = \{ P \mid P\mathbf{1} = r, \, P^T\mathbf{1} = c \}$ is the set of valid transport plans with marginals $r$ and $c$ (both uniform distributions in our case), $\lambda$ is the regularization strength, and $H(P) = -\sum_{ij} P_{ij} \log P_{ij}$ is the entropy of the transport plan.
The Sinkhorn algorithm provides an efficient iterative solution by alternating between row and column normalization. Starting with the kernel matrix $K = \exp\left(\frac{S}{\lambda}\right)$, we iterate:

\begin{equation}
u^{(t+1)} = \frac{\mathbf{1}}{Kv^{(t)}}
\end{equation}

\begin{equation}
v^{(t+1)} = \frac{\mathbf{1}}{K^T u^{(t+1)}}
\end{equation}

The final transport plan is computed as:
\begin{equation}
P = \text{diag}(u) \cdot K \cdot \text{diag}(v)
\end{equation}
In our experiments, we use $\lambda = 5.0$ , striking a balance between computational efficiency and solution quality. The larger $λ$ value ensures numerical stability by preventing extreme values in the exponential computation.
The key advantage of Sinkhorn regularization is that its inherent noising suppression capability. The doubly-stochastic constraint forces the attention weights to be distributed across multiple keys, so that any noise value can be prevented from dominating the attention even if it produces an high score. Additionally, the entropy regularization term penalizes concentrated attention distributions, promoting a smoother attention pattern that is less sensitive to individual outliers.
The OT effect on attention is illustrated in Fig 3. 
\subsection{Adaptive Score Fusion}
We combine the base attention scores $S$, the convolution-enhanced scores $S_{\text{conv}}$, and the OT-regularized scores $P$ through a learnable weighted fusion:
\begin{equation}
S_{\text{final}} = \alpha \cdot S + \beta \cdot S_{\text{conv}} + \gamma \cdot P
\end{equation}
where $\alpha$, $\beta$, $\gamma$ are learnable parameters initialized to 0.7, 0.2, and 0.1, respectively. We normalize the weights using softmax for ensuring stable training and prevent any component from dominating:
\begin{equation}
[\alpha', \beta', \gamma'] = \text{softmax}([\alpha, \beta, \gamma])
\end{equation}
The initial weight distribution reflects our design idea: the base attention scores provide the main signal, the MSC enhancement offers complementary local structure information, and the OT regularization is used to keep stable and suppress the noise. During training, the model can adaptively adjust these weights based on the characteristics of the dataset.

\subsection{Numerical Stability Considerations}
Real-world time-series data often contains extreme values, which can lead to numerical instability in attention computation. We use two component-class methods (score for pruning and Sinkhorn kernel stabilization) to safeguard our experiments:

Score Pruning: The base attention scores are processed using a tanh activation function and scaled to the range $[-8, 8]$ to prevent extreme values from causing overflow in subsequent exponential operations.

Sinkhorn Kernel Stabilization: Before computing the kernel matrix K, we subtract the maximum value and prune the input values to the range $[-50, 50]$. And we clamp intermediate results during iteration to prevent division by zero.

\subsection{Scope of Applicability}
MSC-OT is specifically designed for the inverted embedding method in MTS forecasting. The effectiveness of our method comes from its ability to compensate for the loss of temporal dependency information caused by inverted embedding methods. When the MSC-OT method is applied to traditional time series embedding methods (where each time step of all variables is treated as an embedded label), its advantages are reduced.

\begin{table*}[htbp]
\caption{Multivariate forecasting results with prediction lengths S in [96, 192,336,720] for datasets in [ECL, ETT, Exchange, Traffic, Weather] and the results are averaged from all prediction lengths. Avg means further averaged by 4 subsets of ETT. }
\begin{center}
\resizebox{\textwidth}{!}{
\begin{tabular}{|c|cc|cc|cc|cc|cc|cc|cc|cc|cc|}
\hline
\textbf{Models} & \multicolumn{2}{c|}{\textbf{MSC-OT(Ours)}} & \multicolumn{2}{c|}{\textbf{iTransformer}} & \multicolumn{2}{c|}{\textbf{RLinear}} & \multicolumn{2}{c|}{\textbf{PatchTST}} & \multicolumn{2}{c|}{\textbf{Crossformer}} & \multicolumn{2}{c|}{\textbf{TimesNet}} & \multicolumn{2}{c|}{\textbf{SCINet}} & \multicolumn{2}{c|}{\textbf{FEDformer}} & \multicolumn{2}{c|}{\textbf{Autoformer}} \\
\hline
\textbf{Metric} & \textbf{\textit{MSE}} & \textbf{\textit{MAE}} & \textbf{\textit{MSE}} & \textbf{\textit{MAE}} & \textbf{\textit{MSE}} & \textbf{\textit{MAE}} & \textbf{\textit{MSE}} & \textbf{\textit{MAE}} & \textbf{\textit{MSE}} & \textbf{\textit{MAE}} & \textbf{\textit{MSE}} & \textbf{\textit{MAE}} & \textbf{\textit{MSE}} & \textbf{\textit{MAE}} & \textbf{\textit{MSE}} & \textbf{\textit{MAE}} & \textbf{\textit{MSE}} & \textbf{\textit{MAE}} \\
\hline
ECL & \textcolor{red}{\textbf{0.172}} & \textcolor{red}{\textbf{0.263}} & \textcolor{blue}{\underline{0.178}} & \textcolor{blue}{\underline{0.270}} & 0.219 & 0.298 & 0.205 & 0.290 & 0.244 & 0.334 & 0.192 & 0.295 & 0.268 & 0.365 & 0.214 & 0.327 & 0.227 & 0.338 \\
\hline
ETT (Avg) & 0.385 & 0.401 & 0.383 & 0.399 & \textcolor{red}{\textbf{0.380}} & \textcolor{red}{\textbf{0.392}} & \textcolor{blue}{\underline{0.381}} & \textcolor{blue}{\underline{0.397}} & 0.685 & 0.578 & 0.391 & 0.404 & 0.689 & 0.597 & 0.408 & 0.428 & 0.465 & 0.459 \\
\hline
Exchange & \textcolor{blue}{\underline{0.363}} & \textcolor{red}{\textbf{0.403}} & \textcolor{red}{\textbf{0.360}} & \textcolor{red}{\textbf{0.403}} & 0.378 & 0.417 & 0.367 & \textcolor{blue}{\underline{0.404}} & 0.940 & 0.707 & 0.416 & 0.443 & 0.750 & 0.626 & 0.519 & 0.429 & 0.613 & 0.539 \\
\hline
Traffic & \textcolor{red}{\textbf{0.419}} & \textcolor{red}{\textbf{0.279}} & \textcolor{blue}{\underline{0.428}} & \textcolor{blue}{\underline{0.282}} & 0.626 & 0.378 & 0.481 & 0.304 & 0.550 & 0.304 & 0.620 & 0.336 & 0.804 & 0.509 & 0.610 & 0.376 & 0.628 & 0.379 \\
\hline
Weather & \textcolor{red}{\textbf{0.255}} & \textcolor{red}{\textbf{0.276}} & \textcolor{blue}{\underline{0.258}} & \textcolor{blue}{\underline{0.278}} & 0.272 & 0.291 & 0.259 & 0.281 & 0.259 & 0.315 & 0.259 & 0.287 & 0.292 & 0.363 & 0.309 & 0.360 & 0.338 & 0.382 \\
\hline
\end{tabular}
}
\label{tab1}
\end{center}
\end{table*}

\section{Experiments}

\subsection{Experiments Setup}

We evaluated the MSC-OT model using five widely used benchmark datasets, including ECL, ETT, Exchange, Traffic and Weather, where ETT consists of 4 subsets (ETTh1, ETTh2, ETTm1 and ETTm2). These datasets cover different application domains and exhibit varying characteristics in terms of dimensionality, temporal granularity, and sequence length.

We use two widely used metrics to quantitatively evaluate forecast performance: Mean Squared Error (MSE) and Mean Absolute Error (MAE). MSE penalizes larger forecast errors more severely and is therefore more sensitive to outliers; while MAE provides a more balanced assessment of average forecast accuracy. For both metrics, lower values indicate better forecast performance. We compare MSC-OT with the recent-year MTS models including iTansformer, RLinear, PatchTST, Crossformer, TimesNet, SCINet, FEDformer and Autoformer.

MSC-OT the model is architecture consists of two encoder layers and one decoder layer, with a model dimension of 512, eight attention heads, a feedforward dimension of 2048, and a dropout rate of 0.1. For the core attention module of MSC-OT, we configured multi-scale convolutions with kernel sizes of $3 \times3$ and $5 \times5$ to capture local patterns at different granularities. To ensure training stability, the convolution weights were initialized at a constant value of 0.1. The optimal sinkhorn transport regularization uses an entropy regularization coefficient $\lambda = 5.0$ and performs two matrix scaling iterations, which we found sufficient to obtain a near-optimal birandom distribution while maintaining computational efficiency. The adaptive fusion mechanism initializes the learnable weights as follows: base attention score $\alpha = 0.7$, convolutional enhancement score $\beta = 0.2$, and OT regularization score $\gamma = 0.1$, using softmax normalization to ensure the sum of the weights is 1. The bound sets as $tan{h}$ and the scaling factor sets as 8.0, that constrains the attention score within a reasonable range, cropping the score to the range [-20.0, 20.0] before OT calculation, and using a safety index limited to the range [-50, 50]. The model is optimized using the Adam optimizer with a learning rate of $1 \times 10^{-4}$, a batch size of 32, and early stopping with a patience value of 3 epochs. The input sequence length is fixed at 96, and prediction ranges of 96, 192, 336, and 720 are used for long-term prediction evaluation. All experiments are repeated 3 times with different random seeds, and the average performance is reported.

\subsection{Main Results}
We present comprehensive multivariate prediction results for prediction timeframes ranging from 96 to 720 time steps, enabling a comprehensive evaluation of both short-term and long-term prediction scenarios.

The experiments show that MSC-OT has the best MSE result on three datasets including ECL, Traffic and Weather, and has suboptimal result of MSE on Exchange. Meanwhile, the MSC-OT has best MAE result on four datasets except for ETT. Experimental results show that MSC-OT outperforms transformer-based multivariate time series models developed in recent years. The result is illustrated in Table 1.

Meanwhile, we also obtained the weight results of each component after adaptive weight fusion. This shows that the weights initially set to [0.7, 0.2, 0.1], after adaptive learning, generally account for about 50\% of the MSC-OT weights across all datasets, indicating that they do indeed have a significant impact on the Attention mechanism and have an optimization effect on datasets of different lengths. Table 2 shows the detailed results of weights of different lengths across the four datasets. Fig. 4 shows a comparison of the proportions of the three weights after averaging across the datasets.

\begin{table}[htbp]
\caption{MSC-OT weighting results on 4 Datasets in different prediction lengths and average results}

\centering
\begin{tabular}{ccccc}
\toprule
Dataset & S lengths & $\alpha$ (Base) & $\beta$ (Conv) & $\gamma$(OT) \\
\midrule
\multirow{4}{*}{Weather} & 96   & 0.505468 & 0.263824 & 0.230708 \\
                          & 192  & 0.502627 & 0.268439 & 0.228934 \\
                          & 336  & 0.499995 & 0.269652 & 0.230353 \\
                          & 720  & 0.496565 & 0.272043 & 0.231392 \\
                          
                          & AVG  & \textbf{0.50116375} & \textbf{0.2684895} & \textbf{0.23034675} \\
\midrule
\multirow{4}{*}{Exchange} & 96   & 0.480226 & 0.273616 & 0.246159 \\
                          & 192  & 0.482446 & 0.273326 & 0.244228 \\
                          & 336  & 0.47198  & 0.277189 & 0.250832 \\
                          & 720  & 0.483815 & 0.272322 & 0.243863 \\
                          
                          & AVG  & \textbf{0.47961675} & \textbf{0.27411325} & \textbf{0.2462705} \\
\midrule
\multirow{4}{*}{Traffic}  & 96   & 0.62072  & 0.278767 & 0.100513 \\
                          & 192  & 0.632647 & 0.267851 & 0.099502 \\
                          & 336  & 0.632168 & 0.263162 & 0.10467  \\
                          & 720  & 0.632136 & 0.261204 & 0.10666  \\
                          
                          & AVG  & \textbf{0.62941775} & \textbf{0.267746} & \textbf{0.10283625} \\
\midrule
\multirow{4}{*}{ECL}      & 96   & 0.420867 & 0.406031 & 0.173101 \\
                          & 192  & 0.443632 & 0.398301 & 0.158067 \\
                          & 336  & 0.431988 & 0.41054  & 0.157472 \\
                          & 720  & 0.405247 & 0.417    & 0.177753 \\
                         
                          & AVG  & \textbf{0.4254335} & \textbf{0.407968} & \textbf{0.16659825} \\
\bottomrule
\end{tabular}
\end{table}

\begin{figure}[htbp]
\centerline{\includegraphics[width=0.5\textwidth]{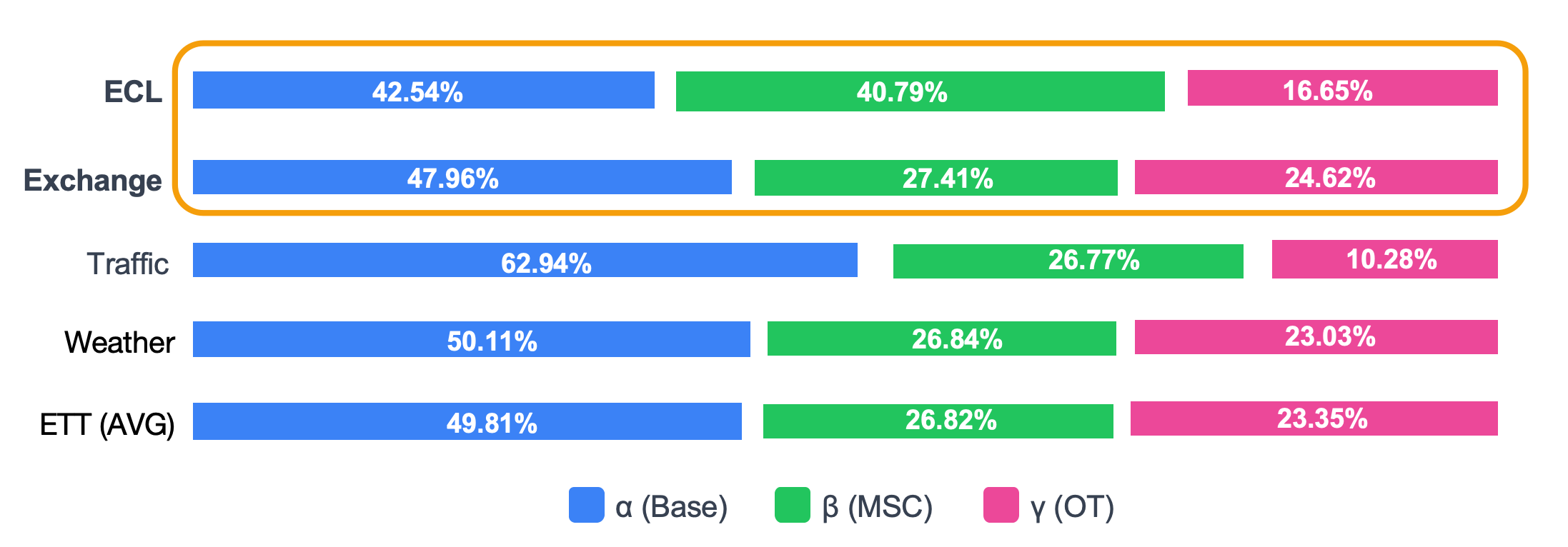}}
\caption{The average weighting results in different datasets. The result shows that MSC-OT has an obvious effect, where ECL and Exchange showing  $\beta + \gamma$ clearly exceeding 50\% means that balanced attention more critical}
\label{fig}
\end{figure}

\subsection{Ablation Experiments}
The ablation experiments are conducted to validate the effectiveness of each proposed component and understand their individual and combined contributions. The experiments systematically separated the effects of multi-scale convolutional enhancement, Sinkhorn optimal transport regularization, and their synergistic combination.

Sinkhorn Optimal Transport Regularization Analysis. We analyze the OT regularization component by comparing variants with and without the Sinkhorn algorithm. Excluding OT regularization leads to a measurable performance degradation, especially on high-dimensional datasets. The dual stochastic attention distribution enforced by Sinkhorn iterations ensures that each variable contributes information and receives information from other variables in a balanced manner.  OT regularization effectively mitigates this problem by guaranteeing that the row and column sums of the attention matrix converge to a uniform distribution. Furthermore, as described in our methodology, the OT formula suppresses the influence of anomalous noise by treating attention computation as an optimal transmission problem, where extreme values are naturally suppressed through an entropy regularization term controlled by $\lambda$.

\begin{figure}[htbp]
\centerline{\includegraphics[width=0.5\textwidth]{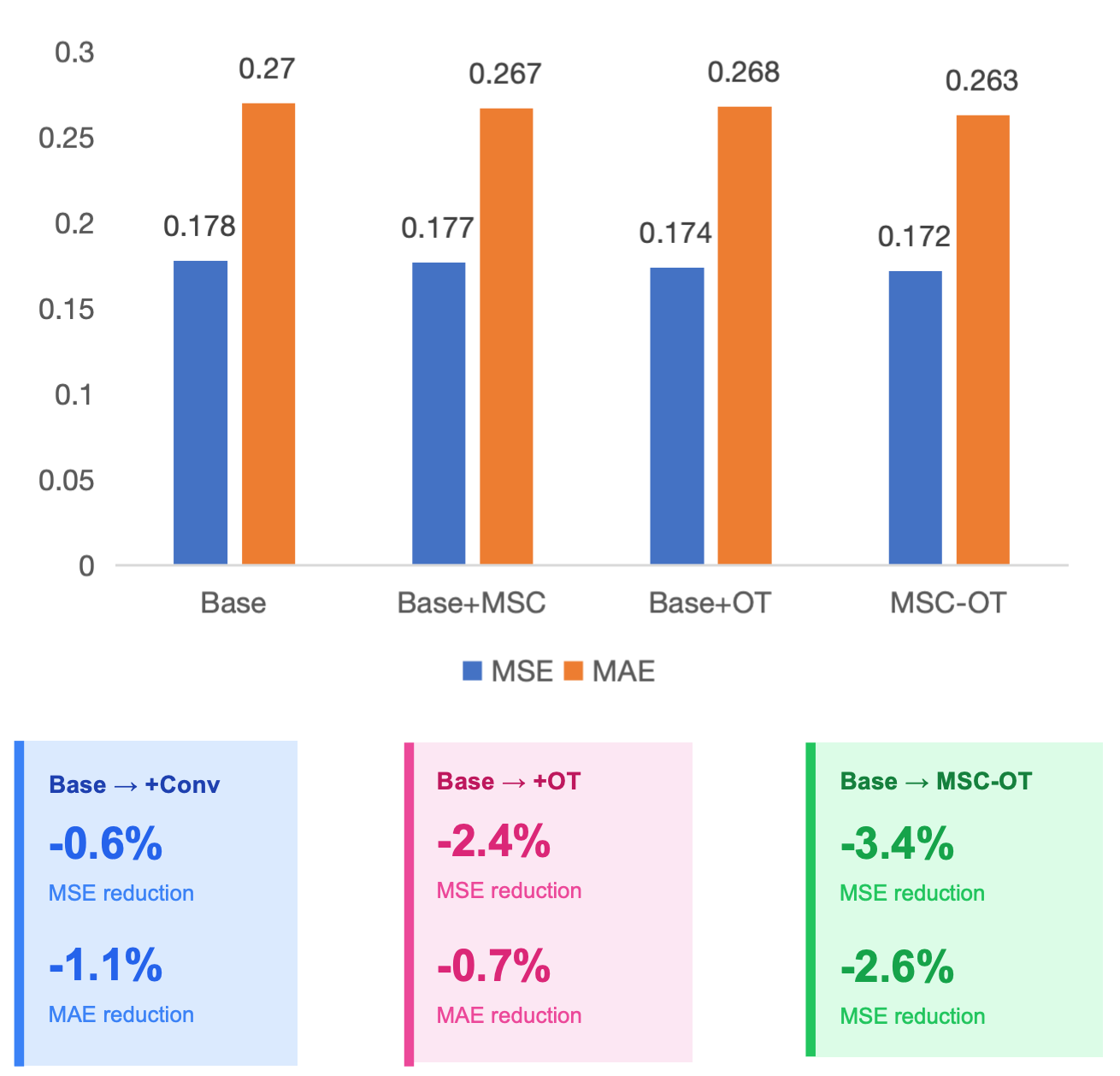}}
\caption{The ablation experiment based on ECL data shows that the combination of multi-scale convolution and optimal transport has a better optimization effect. }
\label{fig}
\end{figure}

Synergistic Effect of Combined Components. Ablation experiments show that multi-scale convolution and Sinkhorn OT regularization have a synergistic effect, instead of any independent one. The ablation experiment result shows a complementary relationship: multi-scale convolution captures local structural patterns and OT regularization ensures a balance in information propagation, so that optimizing the global distributed attention. This combined mechanism allows MSC-OT to simultaneously capture local patterns and balance global information, achieving robust and accurate predictions. The ablation experiment result on ECL is illustrated in Fig 5. 

Adaptive Fusion Strategy Analysis. We further investigated the adaptive fusion mechanism by comparing the learned weights with fixed equal weights ($\alpha=\beta=\gamma{}=\frac{1}{3}$). The results show that the learnable fusion strategy consistently outperforms fixed weights, indicating that the optimal combination of base attention, convolutional augmentation scores, and OT regularization scores varies across datasets and is best determined through end-to-end learning. This adaptive behavior validates our design using learnable fusion weights with softmax normalization.

We investigate the sensitivity of MSC-OT to key hyperparameters, focusing on the design choices reflected in our implementation. OT Regularization Strength ($\lambda$). The entropy regularization coefficient $\lambda$ controls the smoothness of the optimal transport plan. We evaluate $\lambda$ values ranging from 1.0 to 10.0. Smaller values ($ \lambda < 3.0$) lead to numerical instability. Larger values ($\lambda > 8.0$) over-smooth the attention distribution, reducing the model's ability to discriminate between important and less relevant variate relationships. The setting $\lambda$=5.0, as implemented in our model, achieves the optimal trade-off by ensuring stable computation while preserving discriminative attention patterns. This value keeps the exponential inputs within a numerically safe range while allowing sufficient variation in the resulting transport plan.

Sinkhorn Iteration Count. Our ablation experiments result that set iterations from 1 to 5 shows that the iterations count is not very significant to improve the performance. After iterations arrived 2, the MSE and MAE have little been get better result any more. Adding more iterations will have a negligible effect on improvement, but will significantly increase the computational load, showing a linear growth. This finding aligns with theoretical results showing that Sinkhorn iterations converge rapidly for well-conditioned problems. 
Fig 6 shows the variation trends of MSE and MAE based on different iteration numbers.

\begin{figure}[htbp]
\centerline{\includegraphics[width=0.5\textwidth]{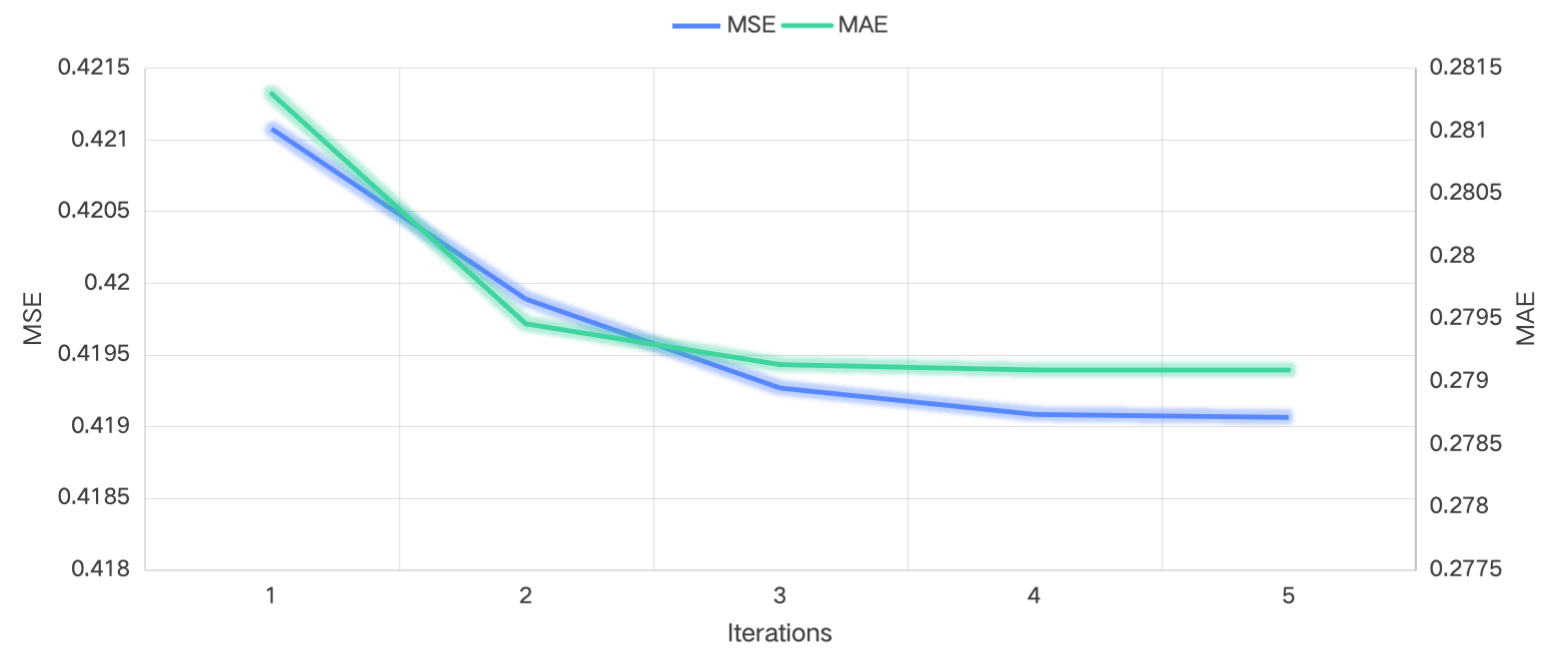}}
\caption{The ablation experiment based on Traffic data shows that iteration T=2 provides an excellent trade-off between approximation quality and computational efficiency. }
\label{fig}
\end{figure}

\section{Summary}
Sufficient experimental evaluation shows that MSC-OT achieves state-of-the-art performance across most of datasets for both short-term and long-term multivariate time series forecasting. Ablation studies confirm that multi-scale convolution enhancement and Sinkhorn optimal transport regularization each provide distinct and complementary benefits: convolution captures local structural patterns at multiple granularities, while OT regularization ensures balanced cross-variate information and suppresses the noise. The adaptive fusion mechanism learns to optimally combine these components based on dataset characteristics. MSC-OT exhibits well robustness to outlier noise, validating the theoretical motivation of our MSC-OT approach,  establishing MSC-OT as a practical and effective solution for real-world multivariate time series forecasting applications and exploring a new  methodological path to optimize attention based on optimal transport.

\bibliographystyle{IEEEtran}
\bibliography{ref.bib}

@article{vaswani2017attention,
  title={Attention is all you need},
  author={Vaswani, Ashish and Shazeer, Noam and Parmar, Niki and Uszkoreit, Jakob and Jones, Llion and Gomez, Aidan N and Kaiser, {\L}ukasz and Polosukhin, Illia},
  journal={Advances in neural information processing systems},
  volume={30},
  year={2017}
}

@article{brown2020language,
  title={Language models are few-shot learners},
  author={Brown, Tom and Mann, Benjamin and Ryder, Nick and Subbiah, Melanie and Kaplan, Jared D and Dhariwal, Prafulla and Neelakantan, Arvind and Shyam, Pranav and Sastry, Girish and Askell, Amanda and others},
  journal={Advances in neural information processing systems},
  volume={33},
  pages={1877--1901},
  year={2020}
}

@article{cuturi2013sinkhorn,
  title={Sinkhorn distances: Lightspeed computation of optimal transport},
  author={Cuturi, Marco},
  journal={Advances in neural information processing systems},
  volume={26},
  year={2013}
}

@article{dosovitskiy2020image,
  title={An image is worth 16x16 words: Transformers for image recognition at scale},
  author={Dosovitskiy, Alexey},
  journal={arXiv preprint arXiv:2010.11929},
  year={2020}
}

@inproceedings{lai2018modeling,
  title={Modeling long-and short-term temporal patterns with deep neural networks},
  author={Lai, Guokun and Chang, Wei-Cheng and Yang, Yiming and Liu, Hanxiao},
  booktitle={The 41st international ACM SIGIR conference on research \& development in information retrieval},
  pages={95--104},
  year={2018}
}

@article{li2017diffusion,
  title={Diffusion convolutional recurrent neural network: Data-driven traffic forecasting},
  author={Li, Yaguang and Yu, Rose and Shahabi, Cyrus and Liu, Yan},
  journal={arXiv preprint arXiv:1707.01926},
  year={2017}
}

@inproceedings{liu2022pyraformer,
title={Pyraformer: Low-Complexity Pyramidal Attention for Long-Range Time Series Modeling and Forecasting},
author={Liu, Shizhan and Yu, Hang and Liao, Cong and Li, Jianguo and Lin, Weiyao and Liu, Alex X and Dustdar, Schahram},
booktitle={International Conference on Learning Representations},
year={2022}
}

@article{liu2023itransformer,
  title={iTransformer: Inverted Transformers Are Effective for Time Series Forecasting},
  author={Liu, Yong and Hu, Tengge and Zhang, Haoran and Wu, Haixu and Wang, Shiyu and Ma, Lintao and Long, Mingsheng},
  journal={arXiv preprint arXiv:2310.06625},
  year={2023}
}

@article{mena2018learning,
  title={Learning latent permutations with gumbel-sinkhorn networks},
  author={Mena, Gonzalo and Belanger, David and Linderman, Scott and Snoek, Jasper},
  journal={arXiv preprint arXiv:1802.08665},
  year={2018}
}

@article{nie2022time,
  title={A time series is worth 64 words: Long-term forecasting with transformers. arXiv 2022},
  author={Nie, Yuqi and Nguyen, Nam H and Sinthong, Phanwadee and Kalagnanam, Jayant},
  journal={arXiv preprint arXiv:2211.14730},
  year={2022}
}

@article{rangapuram2018deep,
  title={Deep state space models for time series forecasting},
  author={Rangapuram, Syama Sundar and Seeger, Matthias W and Gasthaus, Jan and Stella, Lorenzo and Wang, Yuyang and Januschowski, Tim},
  journal={Advances in neural information processing systems},
  volume={31},
  year={2018}
}

@article{salinas2020deepar,
  title={DeepAR: Probabilistic forecasting with autoregressive recurrent networks},
  author={Salinas, David and Flunkert, Valentin and Gasthaus, Jan and Januschowski, Tim},
  journal={International journal of forecasting},
  volume={36},
  number={3},
  pages={1181--1191},
  year={2020},
  publisher={Elsevier}
}

@inproceedings{sander2022sinkformers,
  title={Sinkformers: Transformers with doubly stochastic attention},
  author={Sander, Michael E and Ablin, Pierre and Blondel, Mathieu and Peyr{\'e}, Gabriel},
  booktitle={International Conference on Artificial Intelligence and Statistics},
  pages={3515--3530},
  year={2022},
  organization={PMLR}
}

@inproceedings{tay2020sparse,
  title={Sparse sinkhorn attention},
  author={Tay, Yi and Bahri, Dara and Yang, Liu and Metzler, Donald and Juan, Da-Cheng},
  booktitle={International conference on machine learning},
  pages={9438--9447},
  year={2020},
  organization={PMLR}
}

@article{wu2022timesnet,
  title={Timesnet: Temporal 2d-variation modeling for general time series analysis},
  author={Wu, Haixu and Hu, Tengge and Liu, Yong and Zhou, Hang and Wang, Jianmin and Long, Mingsheng},
  journal={arXiv preprint arXiv:2210.02186},
  year={2022}
}

@article{wu2021autoformer,
  title={Autoformer: Decomposition transformers with auto-correlation for long-term series forecasting},
  author={Wu, Haixu and Xu, Jiehui and Wang, Jianmin and Long, Mingsheng},
  journal={Advances in neural information processing systems},
  volume={34},
  pages={22419--22430},
  year={2021}
}

@inproceedings{wu2020connecting,
  title={Connecting the dots: Multivariate time series forecasting with graph neural networks},
  author={Wu, Zonghan and Pan, Shirui and Long, Guodong and Jiang, Jing and Chang, Xiaojun and Zhang, Chengqi},
  booktitle={Proceedings of the 26th ACM SIGKDD international conference on knowledge discovery \& data mining},
  pages={753--763},
  year={2020}
}

@inproceedings{zeng2023transformers,
  title={Are transformers effective for time series forecasting?},
  author={Zeng, Ailing and Chen, Muxi and Zhang, Lei and Xu, Qiang},
  booktitle={Proceedings of the AAAI conference on artificial intelligence},
  volume={37},
  number={9},
  pages={11121--11128},
  year={2023}
}

@inproceedings{zhang2023crossformer,
  title={Crossformer: Transformer utilizing cross-dimension dependency for multivariate time series forecasting},
  author={Zhang, Yunhao and Yan, Junchi},
  booktitle={The eleventh international conference on learning representations},
  year={2023}
}

@inproceedings{zhou2021informer,
  title={Informer: Beyond efficient transformer for long sequence time-series forecasting},
  author={Zhou, Haoyi and Zhang, Shanghang and Peng, Jieqi and Zhang, Shuai and Li, Jianxin and Xiong, Hui and Zhang, Wancai},
  booktitle={Proceedings of the AAAI conference on artificial intelligence},
  volume={35},
  number={12},
  pages={11106--11115},
  year={2021}
}

@inproceedings{zhou2022fedformer,
  title={Fedformer: Frequency enhanced decomposed transformer for long-term series forecasting},
  author={Zhou, Tian and Ma, Ziqing and Wen, Qingsong and Wang, Xue and Sun, Liang and Jin, Rong},
  booktitle={International conference on machine learning},
  pages={27268--27286},
  year={2022},
  organization={PMLR}
}

@article{bai2018empirical,
  title={An Empirical Evaluation of Generic Convolutional and Recurrent Networks for Sequence Modeling},
  author={Bai, Shaojie},
  journal={arXiv preprint arXiv:1803.01271},
  year={2018}
}

@article{liu2022scinet,
  title={Scinet: Time series modeling and forecasting with sample convolution and interaction},
  author={Liu, Minhao and Zeng, Ailing and Chen, Muxi and Xu, Zhijian and Lai, Qiuxia and Ma, Lingna and Xu, Qiang},
  journal={Advances in Neural Information Processing Systems},
  volume={35},
  pages={5816--5828},
  year={2022}
}

@inproceedings{wang2023micn,
  title={Micn: Multi-scale local and global context modeling for long-term series forecasting},
  author={Wang, Huiqiang and Peng, Jian and Huang, Feihu and Wang, Jince and Chen, Junhui and Xiao, Yifei},
  booktitle={The eleventh international conference on learning representations},
  year={2023}
}

@article{li2024multi,
  title={Multi-scale dilated convolution network for long-term time series forecasting},
  author={Li, Feifei and Guo, Suhan and Han, Feng and Zhao, Jian and Shen, Furao},
  journal={arXiv preprint arXiv:2405.05499},
  year={2024}
}

@article{wang2024timemixer,
  title={Timemixer: Decomposable multiscale mixing for time series forecasting},
  author={Wang, Shiyu and Wu, Haixu and Shi, Xiaoming and Hu, Tengge and Luo, Huakun and Ma, Lintao and Zhang, James Y and Zhou, Jun},
  journal={arXiv preprint arXiv:2405.14616},
  year={2024}
}

@inproceedings{luo2024moderntcn,
  title={Moderntcn: A modern pure convolution structure for general time series analysis},
  author={Luo, Donghao and Wang, Xue},
  booktitle={The twelfth international conference on learning representations},
  pages={1--43},
  year={2024}
}

@article{xu2024step,
  title={Step-wise distribution alignment guided style prompt tuning for source-free cross-domain few-shot learning},
  author={Xu, Huali and Liu, Li and Liu, Tianpeng and Zhi, Shuaifeng and Sun, Shuzhou and Cheng, Ming-Ming},
  journal={arXiv preprint arXiv:2411.10070},
  year={2024}
}

@article{song2025d2vformer,
  title={D2Vformer: A Flexible Time-Series Prediction Model Based on Time-Position Embedding},
  author={Song, Xiaobao and Wang, Hao and Deng, Liwei and Wang, Dong and Qiu, Hongbo and He, Yuxin and Cao, Wenming and Leung, Chi-Sing},
  journal={IEEE Transactions on Neural Networks and Learning Systems},
  year={2025},
  publisher={IEEE}
}

@inproceedings{liu2025timecma,
  title={{TimeCMA}: Towards LLM-Empowered Multivariate Time Series Forecasting via Cross-Modality Alignment},
  author={Liu, Chenxi and Xu, Qianxiong and Miao, Hao and Yang, Sun and Zhang, Lingzheng and Long, Cheng and Li, Ziyue and Zhao, Rui},
  booktitle={AAAI},
  year={2025}
}

\end{document}